\documentclass{article}

\usepackage[preprint]{neurips_2022}

\usepackage[utf8]{inputenc} 
\usepackage[T1]{fontenc}    
\usepackage{hyperref}       
\usepackage{url}            
\usepackage{booktabs}       
\usepackage{amsfonts}       
\usepackage{nicefrac}       
\usepackage{microtype}      
\usepackage{xcolor}         
\usepackage{bm}

\usepackage{graphicx}
\usepackage{amsmath}
\usepackage{amssymb}
\usepackage{amsthm}
\usepackage{dsfont}
\usepackage{multirow}

\usepackage{xcolor}
\usepackage{tikz}
\usetikzlibrary{shapes}
\usetikzlibrary{arrows, arrows.meta}
\usepackage{bm}
\usepackage{dirtytalk}

\newtheoremstyle{mytheorem}
  {5pt}
  {3pt}
  {\itshape}
  {}
  {\itshape\bfseries}
  {.}
  {.5em}
  {\thmname{#1}\thmnumber{{ }#2}%
   \thmnote{ {\the\thm@notefont(#3)}}}
\makeatother
\theoremstyle{mytheorem}
\newtheorem{definition}{Definition}

\newtheorem{lemma}{Lemma}
\newtheorem{remark}{Remark}
\newtheorem{proposition}{Proposition}

\DeclareMathOperator{\rank}{rank}
\DeclareMathOperator{\nullity}{nullity}

\title{Exact Feature Collisions in Neural Networks}

\author{
  Utku Ozbulak\,\thanks{Corresponding author: \texttt{utku.ozbulak@ugent.be}} \,
  \thanks{Department of Electronics and Information Systems, Ghent University, Belgium} \,
  \thanks{Center for Biosystems and Biotech Data Science, Ghent University Global Campus, Republic of Korea}
   \And Manvel Gasparyan  \footnotemark[3] \,   \thanks{Department of Data Analysis and Mathematical Modelling, Ghent University, Belgium}
   \And Shodhan Rao \footnotemark[3] \, \footnotemark[4]
   \AND Wesley De Neve \footnotemark[2] \,  \footnotemark[3] 
   \And Arnout Van Messem \thanks{Department of Mathematics, University of Li\`ege, Belgium}
}
\begin{document}

\maketitle

\begin{abstract}
Predictions made by deep neural networks were shown to be highly sensitive to small changes made in the input space where such maliciously crafted data points containing small perturbations are being referred to as adversarial examples. On the other hand, recent research suggests that the same networks can also be extremely insensitive to changes of large magnitude, where predictions of two largely different data points can be mapped to approximately the same output. In such cases, features of two data points are said to \textit{approximately collide}, thus leading to the largely similar predictions. Our results improve and extend the work of~\citet{li2019approximate}, laying out theoretical grounds for the data points that have colluding features from the perspective of weights of neural networks, revealing that neural networks not only suffer from features that approximately collide but also suffer from features that \textit{exactly collide}. We identify the necessary conditions for the existence of such scenarios, hereby investigating a large number of DNNs that have been used to solve various computer vision problems. Furthermore, we propose the Null-space search, a numerical approach that does not rely on heuristics, to create data points with colliding features for any input and for any task, including, but not limited to, classification, localization, and segmentation.
\end{abstract}

\section{Introduction}\label{sec:intro}

Since the inception of deep neural networks (DNNs) with AlexNet~\citep{Alexnet}, computer vision problems involving classification, segmentation, and localization found quick-to-adopt solutions~\citep{resnet,mask_rcnn,mobilenet_v3}. Shortly after the wide-range adoption of DNNs in computer vision, other domains such as natural language processing~\citep{devlin2018bert}, sequence analysis~\citep{zuallaert2018splicerover}, and radar-based detection~\citep{vandersmissen2019indoor} embraced the same networks and utilized them to solve a wide-range of complex problems.

Much to the surprise of the research community, features learned by DNNs were found to be easily abusable by adversarial examples~\citep{biggio2013evasion,Goodfellow-expharnessing}, malicious data points created with the sole purpose of misleading machine learning models. Such adversarial examples are created by adding a small perturbation to an image, which then changes the prediction of that image drastically~\citep{LBFGS}, meaning that small changes in inputs may lead to large changes in outputs (i.e., predictions). Although adversarial examples are shown to be a threat in other domains, the vision domain in particular suffers greatly from this phenomenon due to the exercised perturbation often being invisible to the naked eye. As a result, mission-critical tasks such as self-driving cars~\citep{self_driving_adversarial} and medical image diagnosis~\citep{ozbulak2019impact}, which typically take advantage of the predictive power of DNNs, suffer greatly from this vulnerability.

After the discovery of adversarial examples~\citep{Goodfellow-expharnessing}, substantial research efforts were spent to find a solution to this security flaw~\citep{CarliniTeam_2020_defense_evaluation}. Among those research efforts, a subset focused their effort to demystify the black-box nature of neural networks and to improve the understanding of the feature space~\citep{ilyas2019adversarial}. While the research on the topic of adversarial examples has shown that neural networks are extremely sensitive to small amounts of (adversarial) perturbations, the work of~\citet{jacobsen2018excessive} and~\cite{li2019approximate} have revealed the that same neural networks are also impartial to certain other types of perturbations. Particularly,~\citet{li2019approximate} have shown that neural networks that contain ReLU activations suffer from what they called \textit{approximate feature collisions} where multiple (dissimilar) data points may have similar feature maps, leading to similar predictions. 

In this work, we investigate the phenomenon of colliding features in order to improve the understanding of the neural network feature space. In particular, we improve and extend the work of~\citet{li2019approximate} to trainable weights and reveal the existence of data points with \textit{exactly} colliding features, irrespective of the selected activation function. Based on our findings, we propose the Null-space search, a novel numerical method that leverages the null-space of weights in creating colliding data points. Finally, expanding on the angle of adversarial vulnerability, we proceed to show the existence of adversarial examples with colliding features, which exacerbates the adversarial risk associated with DNNs.

\section{Related Work}

Iterative generation of adversarial examples has been a topic that gained traction in recent years~\citep{dong2018boosting,wang2020transferable_cross} where many recently-proposed iterative attacks follow in the footsteps of~\cite{LBFGS} and~\cite{IFGS}. The work of~\cite{sabour2015adversarial} on the other hand approached the topic of adversarial examples from a different angle, where they tried to create adversarial examples that come with particular and pre-defined feature representations. Such approaches were employed to fool the interpretability of DNNs~\citep{subramanya2019fooling}. This line of work also led to the discovery of invariance-based adversarial examples~\citep{jacobsen2018excessive}, where these adversarial examples not only come with limited perturbation, but also come with specified predictions. Such adversarial examples comes with an identical (or almost identical) prediction compared to a data point in the dataset, thus making it impossible to differentiate from that genuine data point based on the output of the model. Invariance-based adversarial examples were used in order to circumvent adversarial defenses~\citep{hosseini2019are_odds_odd,mimic_fool,CarliniTeam_2020_defense_evaluation}, understand and manipulate feature representations~\citep{subramanya2019fooling}, and to understand trade-offs between invariance and sensitivity to adversarial perturbations~\cite{tramer2020fundamental}.

Another line of work, and the one on which our paper is mostly based, is the work of~\cite{li2019approximate} where approximate feature collisions based on activation functions were discovered. This work mainly takes advantage of the shortcomings of DNN activation functions, and in particular, rectifiers~\citep{ReLU}. These shortcomings are also thoroughly discussed in the work of~\cite{hein2019relu}.

\section{Preliminaries}
\label{Preliminaries}

Let $\mathcal{D}= \{\bm{x}_n \,|\, n=1, \ldots, N \}$ be a dataset containing $N$ data points. We denote by $\bm{x}_n \in \mathbb{R}^q$ and $y_n \in \{1, \ldots, M\}$ a single data point in $\mathcal{D}$ and its categorical information, respectively. In this setting, let $g : \mathbb{R}^q \to \mathbb{R}^M$ denote a neural network classifier that yields a vector of real-valued logit scores for each category. Logits obtained by the usage of a neural network with parameters $\theta$ are represented by $g(\theta, \bm{x}) \in  \mathbb{R}^M$. In this system, the index that contains the largest logit is assigned as the categorical classification for a data point, $G(\theta, \bm{x}_n) = \arg \max_m \big( g(\theta, \bm{x}_n)_m \big)$. If $G(\theta, \bm{x}_n) = y_n$, the data point is correctly classified.

We represent the internal working of a neural network as a composition of layers
\begin{eqnarray}
\label{eq:nn_representation}
g(\theta, \bm{x}) = f^{(t)}(\bm{x}) := f_t \circ f_{t-1} \circ \ldots \circ f_1(\bm{x}) \,,
\end{eqnarray}
with $f_k(\cdot)$ and $f^{(k)}(\cdot)$ indicating a forward pass at the $k^{th}$ layer and a forward pass until the $k^{th}$ layer, respectively. Each of the layers can perform any of the commonly used operations in standard neural network architectures, such as convolution, attention, activation, and pooling~\citep{lecun1998gradient,ReLU,attention}.

\section{Feature Collisions}
\label{Feature Collision}
For neural networks, feature collisions can be defined as follows.

\begin{definition}
\label{def:any_layer_feature_collision}
(Feature collisions) Given two data points $\bm{x}_1 \neq \bm{x}_2$ and a neural network with parameters $\theta$. If $f^{(k)}(\bm{x}_1) = f^{(k)}(\bm{x}_2)$ for a layer $k$ then features of $\bm{x}_1$ and $\bm{x}_2$ collide on the $k^{th}$ layer.
\end{definition}

Although colliding features have a number of implications, the most important one is having two different inputs being mapped to the same output. Here, we would like to make the following remark about inference using neural networks.
\begin{remark}
\label{rem:nn_layers_forward}
(Inference with neural networks) Since the layer operations for a neural network described in Eq.~(\ref{eq:nn_representation}) are deterministic during inference time, at the moment of inference, we have that if \mbox{$f^{(k)}(\bm{x}_1) = f^{(k)}(\bm{x}_2)$}, then  \mbox{$g(\theta, \bm{x}_1) = g(\theta, \bm{x}_2)$}.
\end{remark}

Remark~\ref{rem:nn_layers_forward} is directly related to the property of injectiveness. This implies that if a neural network has data points with colliding features, it is not injective, and vice versa. Then, given a neural network $\theta$ of the form of Eq.~(\ref{eq:nn_representation}), if 
\begin{eqnarray}
\label{eq:inv_equivalence}
\min_{\bm{x}_1 \neq \bm{x}_2 \neq \bm{0}}|| g(\theta, \bm{x}_1) - g(\theta, \bm{x}_2) || = 0 
\end{eqnarray}
holds, that neural network is not injective. Based on Remark~\ref{rem:nn_layers_forward}, we know that if one of the layers in $\theta$ is not injective, then $g$ itself is not injective. In this context, a non-injective transformation indicates that the domain of the transformation is larger than its image, as such, there will always be at least two different elements in the domain that will be mapped to the same element.

\begin{proposition}
\label{prop:non-injective}
For any neural network represented by $\theta$, Eq.~(\ref{eq:inv_equivalence}) holds if $g(\theta, \bm{x})$ is non-injective with respect to its second argument.
\end{proposition}

In order to investigate the plausibility of Proposition~\ref{prop:non-injective}, we need to determine the conditions for which Eq.~(\ref{eq:inv_equivalence}) holds.

\citet{li2019approximate} investigated feature collisions and the validity of Proposition~\ref{prop:non-injective} based on activation functions and defined colliding features based on activations as follows:

\begin{definition}
\label{def:act_feature_collision}
(Feature collisions on activations) Given two data points $\bm{x}_1 \neq \bm{x}_2$ and a neural network with parameters $\theta$. If $\alpha(f^{(k-1)}(\bm{x}_1)) = \alpha(f^{(k-1)}(\bm{x}_2))$ for a layer $k$ and activation function $\alpha$, then features of $\bm{x}_1$ and $\bm{x}_2$ collide on the $k^{th}$ layer.
\end{definition}

In their work, \citet{li2019approximate} scrutinized ReLU activation. ReLU, as defined by $\text{ReLU}(z) = \max\{0, z\}$, is indeed non-injective: by definition, $\forall z_1 \neq z_2 \in \mathbb{R}_{-}$ holds that $\text{ReLU}(z_1) = \text{ReLU}(z_2) = 0$. 

Let us define the features of $\bm{x}$ at the $k^{th}$ layer as $\bm{l}^{(x)} = \text{ReLU}(f^{(k-1)}(\bm{x}))$ based on Definition~\ref{def:act_feature_collision} and ReLU activation. Two data points $\bm{x}_{1}$ and $\bm{x}_{2}$ have a feature collision if $\bm{l}_{+}^{(x_1)} = \bm{l}^{(x_2)}_{+}$, where $\bm{l}_{+}$ denotes non-negative elements of $\bm{l}$, since all the negative elements will be mapped to $0$ due to the ReLU activation. Because of the property highlighted above, thus far, the creation of data points with colliding features for DNNs containing such non-injective activation functions has been possible~\citep{jacobsen2018excessive}. 

For other (injective) activation functions such as sigmoid or tanh, \citet{li2019approximate} argue that the property of saturation (e.g., $\lim_{x\to \infty} \text{sigmoid}(x) = 1$) may lead to \textit{approximate collisions}. As a result, the observations regarding the colliding features made in the aforementioned work are mostly applicable to ReLU networks and do not generalize easily to others other networks that do not have this type of activation.

Another commonly used building block of neural networks are the pooling layers. These layers are, more-often-than-not, employed to reduce the dimension of the feature maps, thus reducing the number of parameters required to learn representative features. When investigating feature collisions, we can easily take advantage of the dimensionality of the feature space and show the existence of colliding features.

\begin{remark}
\label{rem:pooling}
(Feature collisions on pooling) Let $\gamma:\mathbb{R}^{q}\to\mathbb{R}^{p}$ be a pooling operation. If $q > p$ then, by definition, $\exists \bm{z}_1 \neq \bm{z}_2 \in \mathbb{R}^{q}$ such that $\gamma(\bm{z}_1) = \gamma(\bm{z}_2)$.
\end{remark}

Let us consider max-pooling, which is the most commonly used pooling operation, to demonstrate the feasibility of Remark~\ref{rem:pooling}. Suppose $\gamma:\mathbb{R}^{3}\to\mathbb{R}$ is a $1\times3$ max-pooling operation, and $\bm{z}=[z_{1}, z_{2}, z_{3}]$ an input to this operation where $\min(\bm{z}) = z_{1}$. If we replace $z_{1}$ by $z_{\star} \in \mathbb{R}$ such that $z_{\star} < z_{1}$, and subsequently write $\bm{z}_\star = [z_{\star}, z_{2}, z_{3}]$, then any input created by this approach will result in $\gamma(\bm{z}) = \gamma(\bm{z}_{\star})$.

Although ReLU activations and pooling functions used to be popular building blocks of neural networks, the former has been increasingly replaced with other types of activation functions and the latter has seen reduced use in recently proposed architectures~\cite{hendrycks2016gelu,touvron2020training,dosovitskiy2021an}. Different from those building blocks, trainable parameters are perpetual in DNN architectures, either in the form of convolutional or fully-connected layers. As such, different from previous research discussed thus far, we focus our attention on trainable weights and analyze their properties for the colliding features, independent of the data used for training as well as activation functions or pooling present in the model.

\subsection{Feature collision on weights}
\label{Feature_collision_on_weights}

Based on a neural network of the form of~Eq.~(\ref{eq:nn_representation}), we define the $k^{th}$ layer as $f^{(k)}(\bm{x}) = \bm{x}^{\top}\bm{W}^{(k)} + \bm{b}^{(k)}$.
We then define feature collisions on, we define feature collisions on trainable layers as follows.

\begin{definition}
\label{def:weight_feature_collision}
(Feature collisions on weights) Given two data points $\bm{x}_1  \neq \bm{x}_2$ and a trainable weight at $k^{th}$ layer $f^{(k)}(\bm{x})$. If the we have $f^{(k)}(\bm{x}_1) = f^{(k)}(\bm{x}_2)$, the features of $\bm{x}_1$ and $\bm{x}_2$ collide on the $k^{th}$ layer.
\end{definition}

Our analysis starts with fully connected layers, where we make the following observation regarding the eigenvalues of the weights and the feasibility of having colliding features.

\begin{lemma}
\label{lem:linear_layer}
(Feature collision on fully connected layers) Let $\bm{v}_k = f^{(k-1)} (\bm{x}) \in \mathbb{R}^{q}$ be a vector output obtained from composite operations up to the $(k-1)^{th}$ layer and let $\bm{W}_k \in \mathbb{R}^{q\times p}$ be the trainable weights at the same layer. Assume that $f_k(\cdot)$ is a fully connected layer such that $f_k(\bm{v}_k) = \bm{v}_k^{\top} \bm{W}_k$. If zero is an eigenvalue of $\mathcal{W}_k=\bm{W}_k \bm{W}_k^{\top}$, then Eq.~(\ref{eq:inv_equivalence}) holds.
\end{lemma}

\begin{proof}
Having a zero eigenvalue for $\mathcal{W}_k$ implies that $\rank(\mathcal{W}_k) < q$. Since $\rank(\mathcal{W}_k) = \rank(\bm{W}_k^{\top})$, it holds that $\rank(\bm{W}_k^{\top}) < q$. From the rank-nullity theorem, we have
\begin{eqnarray}
\nullity(\bm{W}_k^{\top}) + \rank(\bm{W}_k^{\top}) = q\,,
\end{eqnarray}
where $\nullity(\bm{W}_k^{\top}) = \dim(\ker(\bm{W}_k^{\top})) $. This implies that $\nullity(\bm{W}_k^{\top}) \geq 1$ (i.e., $\ker(\bm{W}_k^{\top})$ is not trivial). Let $\bm{\varphi} \in \ker(\bm{W}_k^{\top})$ be a non-zero basis vector. Then $\bm{W}_k^{\top} \bm{\bm{\varphi}} = \bm{0}$. Transposing both sides, we obtain $ \bm{\varphi}^{\top} \bm{W}_k = \bm{0}^{\top}$. Therefore
\begin{eqnarray}
\label{eq:lem1_6}
(\bm{v}_k + \bm{\varphi})^{\top} \bm{W}_k = \bm{v}_k^{\top} \bm{W}_k + \bm{\varphi}^{\top} \bm{W}_k  = \bm{v}_k^{\top} \bm{W}_k \,.
\end{eqnarray}
Note that Eq.~\eqref{eq:lem1_6} can be written as \mbox{$f_{k}(\bm{v}_k) = f_{k}(\bm{v}_k +\bm{\varphi} )$}. Since $\bm{v}_k + \bm{\varphi} \neq \bm{v}_k$, using Remark~\ref{rem:nn_layers_forward}, we obtain Eq.~\eqref{eq:inv_equivalence}.
\end{proof}

Having a zero eigenvalue for the weight matrix also implies that the kernel of the weight matrix is not trivial (i.e., $\dim(\ker(\bm{W})) = \nullity(\bm{W}) \geq 1 $). As a result, the basis vectors in the kernel, $\bm{\varphi} \in \ker(\bm{W})$, can be used to create colliding features, since $(\bm{v}_k+ \bm{\varphi})^{\top} \bm{W} = \bm{v}_k^{\top} \bm{W}$. The observation made in Lemma~\ref{lem:linear_layer} for fully connected layers can be easily extended to convolutional layers.

\begin{remark}
\label{rem:conv_layer}
(Feature collision on convolutional layers) In the proof of Lemma~\ref{lem:linear_layer}, using the fact that $\ker{(\bm{W})}$ is non-trivial, we constructed a vector $\bm{v}_k + \bm{\varphi} \neq \bm{v}_k \in \mathbb{R}^{q}$ such that Eq.~(\ref{eq:inv_equivalence}) holds. If the $k^{th}$ layer is a convolutional layer such that $f_k(\bm{V}_k) = \bm{V}_k \bm{W}_k$, where $\bm{V}_k \in \mathbb{R}^{l\times q}$, we can construct a non-zero matrix $\bm{P} \in \mathbb{R}^{l\times q}$ such that $(\bm{V}_k + \bm{P}) \bm{W}_k = \bm{V}_k  \bm{W}_k$, by, for example, taking $\bm{P} = [\bm{\varphi} \ldots \bm{\varphi}]^{\top} \in \mathbb{R}^{l\times q}$.
\end{remark}

Using Lemma~\ref{lem:linear_layer} and Remark~\ref{rem:conv_layer}, we show the existence of colliding features for neural networks that have at least a single trainable weight matrix with zero as an eigenvalue. Moreover, having larger weight matrices (i.e., increasing the width of the network) also makes it more likely to have zero as an eigenvalue, thus increasing the likelihood of the architecture at hand to have colliding features.

\begin{table*}[t] 
\centering
\caption{For the models provided in the second column, the number of trainable weights that satisfy Lemma~\ref{lem:linear_layer} ($3^{rd}$ column) and Theorem~\ref{thm:layer_n_m} ($5^{th}$ column) with the percentage of those weights compared to all weights ($4^{th}$ and $6^{th}$ columns) are provided. Moreover, the number of basis vectors in the kernel of all trainable weights ($7^{th}$ column) and the number of basis vectors in only the first trainable weight ($8^{th}$ column) are also given.}
\scriptsize
\begin{tabular}{llcccccc}
\cmidrule[0.25pt]{1-8}
Task & Model & $\nu(\theta)$ & $\displaystyle \frac{1}{n_{\theta}} \nu(\theta)$ & $\mu (\theta)$ & $\displaystyle \frac{1}{n_{\theta}} \mu (\theta)$ & $\displaystyle \sum_{\bm{W}_i \in \theta} \nullity(\bm{W}_i) $ & $\nullity(\bm{W}_1)$\\
\cmidrule[0.5pt]{1-8}
\multirow{10}{*}{\rotatebox[origin=c]{90}{Classification}}  & AlexNet~\citep{Alexnet} & $8$ & $100\%$ & $7$ & $87\%$ & $16,547$ & $299$ \\
~  & Squeezenet~\citep{squeezenet} & $16$ & $61\%$ & $16$ & $61\%$ & $3,392$ & $0$ \\
~  & VGG-16~\citep{VGG} & $15$ & $93\%$ & $14$ & $87\%$ & $53,357$ & $0$ \\
~  & ResNet-50~\citep{resnet} & $36$ & $66\%$ & $33$ & $61\%$ & $40,701$ &$84$  \\
~  & DenseNet-121~\citep{densenet} & $121$ & $100\%$ & $117$ & $96\%$ & $90,388$ & $85$ \\
~  & Inception-V3~\citep{inceptionv3} & $95$ & $96\%$ & $95$ & $96\%$ & $80,938$ & $0$ \\
~  & ViT-Base~\citep{dosovitskiy2021an} & $18$ & $36\%$ & $12$ & $24\%$ & $27,665$ & $13$  \\
~  & ViT-Large~\citep{dosovitskiy2021an} & $46$ & $47\%$ & $25$ & $25\%$ & $73,780$ & $0$  \\
~  & DeiT-Tiny~\citep{touvron2020training} & $15$ & $30\%$ & $13$ & $26\%$ & $7,490$ & $576$  \\
~  & DeiT-Base~\citep{touvron2020training} & $16$ & $32\%$ & $12$ & $24\%$ & $27,653$ & $1$  \\
\cmidrule[0.25pt]{1-8}
\multirow{5}{*}{\rotatebox[origin=c]{90}{Segmentation}}  & UNet~\citep{Unet} & $44$ & $93\%$ & $44$ & $93\%$ & $73,626$ & $91$  \\
~  & Fcn~\citep{Fcn8s} & $39$ & $68\%$ & $36$ & $63\%$ & $67,254$ & $84$  \\
~  & PSPNet~\citep{pspnet} & $38$ & $90\%$ & $38$ & $90\%$ & $61,446$ & $90$  \\
~  & MobileNet-V3~\citep{mobilenet_v3} & $45$ & $ 71\%$ & $42$ & $66\%$ & $110,267$ & $84$  \\
~  & Lite R-ASPP~\citep{mobilenet_v3} & $32$ & $48\%$ & $27$ & $40\%$ & $8,756$ & $11$  \\
\cmidrule[0.25pt]{1-8}
\multirow{4}{*}{\rotatebox[origin=c]{90}{Localization}}  & Mask R-CNN~\citep{mask_rcnn} & $54$ & $73\%$ & $50$ & $68\%$ & $74,677$ & $83$  \\
\rule{0pt}{2.2ex}~  & Faster R-CNN~\citep{faster_rcnn} & $38$ & $52\%$ & $32$ & $43\%$ & $27,366$ & $11$  \\
\rule{0pt}{2.2ex}~  & Keypoint R-CNN~\citep{mask_rcnn} & $57$ & $75\%$ & $53$ & $69\%$ & $97,230$ & $84$  \\
\rule{0pt}{2.2ex} ~  & RetinaNet~\citep{focal_loss_retina_net} & $53$ & $ 74\%$ & $50$ & $70\%$ & $72,845$ & $72$  \\
\cmidrule[0.25pt]{1-8}
\cmidrule[1pt]{1-8}
\end{tabular}
\label{tbl:class_seg_loc_nullities}
\end{table*}

\begin{lemma}
\label{lem:layer_width}
(Layer width and eigenvalues) Consider a matrix of the form $\mathcal{W} = \bm{W}\bm{W}^{\top}$, with the corresponding set of eigenvalues $\bm{\Lambda}$. Let $\bm{W}_e$ be a matrix obtained by adding a new row (similarly, a column) to $\bm{W}$ and denote by $\bm{\Lambda}_e$ the set of eigenvalues of $\mathcal{W}_e = \bm{W}_e\bm{W}_e^{\top}$. Then
\begin{eqnarray}
\min (\bm{\Lambda}) \geq \min (\bm{\Lambda}_e) \,.
\end{eqnarray}
\end{lemma}

\begin{proof}
Assume that a vector $\bm{r}$ is added to $\bm{W}$ as new a row, $\bm{W}_e = 
\begin{bmatrix}
\bm{W}\\
\bm{r}^{\top} \\ 
\end{bmatrix}$.
Since $\mathcal{W}_e = \bm{W}_e\bm{W}_e^{\top}$, we have $\mathcal{W}_e = 
\begin{bmatrix}
\mathcal{W} & \bm{W}\bm{r}\\
\bm{r}^{\top}\bm{W}^{\top} & \bm{r}^{\top}\bm{r}  \\
\end{bmatrix}$.
Note that $\mathcal{W}$ is a principal sub-matrix of $\mathcal{W}_e$. From the Courant–Fischer min-max principle \mbox{(see \cite[Theorem~4.3.6]{horn2012matrix})} it thus follows that \mbox{$\min (\bm{\Lambda}) \geq \min (\bm{\Lambda}_e)$}.
\end{proof}

In Lemma~\ref{lem:linear_layer} and Remark~\ref{rem:conv_layer}, we established the existence of colliding features with respect to eigenvalues of the weights. A special case where this effect occurs is when the dimension of the domain of the linear transformation (i.e., $\bm{x}^\top \bm{W}$) is smaller than the dimension of its range.

\begin{lemma}
\label{thm:layer_n_m}
(Inevitability of colliding features) For all neural networks that contain a trainable weight $\bm{W}_k \in \mathbb{R}^{q\times p}$ such that $p < q$, Eq.~(\ref{eq:inv_equivalence}) holds.
\end{lemma}
\begin{proof}
For any $\bm{W}_k \in \mathbb{R}^{q \times p}$ such that $p<q$ we have $\rank{(\bm{W}_k)} = \rank{(\bm{W}^{\top}_k)} \leq p$.
Recalling the rank-nullity theorem, we have $\nullity(\bm{W}_k^{\top}) \geq 1$. Therefore, as a result of Lemma~\ref{lem:linear_layer}, Eq.~\eqref{eq:inv_equivalence} holds.
\end{proof}

We can generalize the observation made in Lemma~\ref{thm:layer_n_m} to an input-output relationship for all neural networks.

\begin{remark}
\label{rem:input_output_size}
(Input-output dimensions and colliding features) For any neural network represented by $\theta$, if $\dim{(\bm{x})} > \dim{\bm{y}}$, where $\bm{y} = g(\theta,\bm{x})$, then a transformation at one of the layers where the dimension of the domain is larger than the dimension of its range takes place. With that fact in mind, Theorem~\ref{thm:layer_n_m} implies that Eq.~(\ref{eq:inv_equivalence}) holds. 
\end{remark}

Lemma~\ref{thm:layer_n_m} together with Remark~\ref{rem:input_output_size} reveals a dire situation for neural networks: it is inevitable to have colliding features for multiple data points when the neural network at hand has an input size (e.g., image size) larger than the size of the output (e.g., number of classes for classification problems). The comparison of input-output size here depends on the first layer of the network. If the first layer is a convolutional layer, the input size we are referring to is the number of elements in the expanded tensor of the input image, where this expansion (also called the unfolding operation) depends on the parameters of the convolution. In the case of the first layer being a linear one, the input size is simply the number of elements in the input matrix/tensor.

In order to demonstrate the practicality of our observations, in Table~\ref{tbl:class_seg_loc_nullities}, we evaluate the properties of a large number of classification, segmentation, and localization models, including the recently-proposed vision transformer models. Specifically, for all trainable weights of the considered models, we calculate 
\begin{eqnarray}
\nu(\theta) = \displaystyle \sum_{\bm{W}_i \in \theta} \mathds{1}_{\{\nullity(W_i)\}}\,, \qquad
\mu (\theta) = \displaystyle \sum_{\bm{W}_i \in \theta} \mathds{1}_{\{\text{row}(W_i) < \text{col}(W_i)\}}\,,
\end{eqnarray}
the number of trainable weights that have zero as an eigenvalue, as shown in Lemma~\ref{lem:linear_layer} and Remark~\ref{rem:conv_layer}, and the number of weights that perform a linear transformation, as shown in Lemma~\ref{thm:layer_n_m}, respectively. With $\frac{1}{n_\theta}\nu(\theta)$ and $\frac{1}{n_\theta}\mu(\theta)$, we also provide the percentage of weights that fit the aforementioned descriptions compared to all weights.  Moreover, with $\sum_{\bm{W}_i \in \theta} \nullity(\bm{W}_i)$, we provide the number of basis vectors in the kernel of weights satisfying Lemma~\ref{lem:linear_layer} and Remark~\ref{rem:conv_layer}. Based on Table~\ref{tbl:class_seg_loc_nullities}, it is clear that, due to a large number of trainable weighs satisfying the conditions we discussed thus far, all of the models listed in Table~\ref{tbl:class_seg_loc_nullities} indeed suffer from potentially colliding features.

Our observations indicate that it is inevitable to have colliding features when using popular DNNs. In the following section, we discuss methods of finding colliding features.

\begin{figure*}
\centering
\begin{tikzpicture}
\foreach \y in {0.25}
{
\draw[draw=gray] (\y-\y*0.2, -3.75) -- (\y-\y*0.2+0.35, -3.75) -- (\y-\y*0.2+0.35, -2.75) -- (\y-\y*0.2, -2.75) -- (\y-\y*0.2, -3.75);
}
\node[align=left, rotate=90] at (0.25-0.25*0.2+0.175, -3.25) {\scriptsize $\bm{x}$};
\foreach \y in {1, 1.75, 2.5, 5.25}
{
\draw[draw=black] (\y-\y*0.2, -4) -- (\y-\y*0.2+0.35, -4) -- (\y-\y*0.2+0.35, -2.5) -- (\y-\y*0.2, -2.5) -- (\y-\y*0.2, -4);
}

\node[align=left, rotate=90] at (1-1*0.2+0.175, -3.25) {\scriptsize $f_1(\cdot)$};
\node[align=left, rotate=90] at (1.75-1.75*0.2+0.175, -3.25) {\scriptsize $f_2(\cdot)$};
\node[align=left, rotate=90] at (2.5-2.5*0.2+0.175, -3.25) {\scriptsize $f_3(\cdot)$};
\node[align=left, rotate=90] at (5.25-5.25*0.2+0.175, -3.25) {\scriptsize $f_t(\cdot)$};
\foreach \y in {1, 1.75, 2.5, 3.25, 5.25}
{
\draw (\y-\y*0.2, -3.25) -- (\y-\y*0.2 - 0.25, -3.25);
}
\node[align=left] at (3.95-3.95*0.2+0.175, -3.25) {$\ldots$};
\node[align=left] at (2.5, -1.9) {\scriptsize Neural network};
\draw[->] (1-1*0.2+0.175, -4) -- (1-1*0.2+0.175, -4.5);
\node[align=left] at (2.5-1*0.2+0.175, -4.75) {$\ker(\bm{W}_1) = \displaystyle \{\bm{\varphi} \in \mathbb{R}^{n} \,|\, \bm{W}_1\bm{\varphi} = 0\}$};

\draw[->] (5-1*0.2+0.175, -4.75) -- (6-1*0.2+0.175, -4.75);
\node[inner sep=0pt] (c) at (7, -5.05){\includegraphics[width=.1\textwidth]{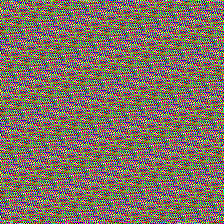}};
\node[align=left] at (7, -6.15) {\scriptsize Perturbation};
\node[align=left] at (7, -6.4) {\scriptsize $\bm{P} = \beta \, [\bm{\varphi} \ldots \bm{\varphi}]$};
\def\xposcirc{7.25}
    \draw[thick,color=red] (7, -4.35) circle (0.1cm);
    \draw[dashed,color=red] (7.1, -4.35) -- (7.5, -3.1);
    \draw[dashed,color=red] (6.9, -4.35) -- (6.5, -3.1);
\node[inner sep=0pt] (c) at (7, -2.45){\frame{\includegraphics[width=.07\textwidth]{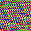}}};
\node[align=center] at (7, -1.15) {\scriptsize Perturbation};
\node[align=center] at (7, -1.4) {\scriptsize (zoomed)};

\node[align=center] at (6.15, -2.52) {\scriptsize $\bm{\varphi}_k$};
\draw[draw=red] (6.45, -2.57) -- (6.45, -2.47) -- (7.55, -2.47) -- (7.55, -2.57) -- (6.45, -2.57);

\node[align=left] at (8.25, -5.05) {$+$};
\node[inner sep=0pt] (c) at (9.5, -5.05){\includegraphics[width=.1\textwidth]{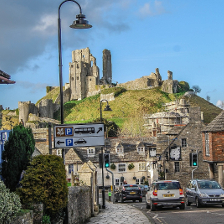}};
\node[align=left] at (9.5, -6.15) {\scriptsize Original};
\node[align=left] at (9.5, -6.4) {\scriptsize image};
\draw[->] (9.5, -4) -- ((9.5, -3.5);
\node[inner sep=0pt] (c) at (9.5, -2.45){\frame{\includegraphics[width=.1\textwidth]{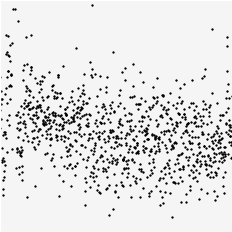}}};
\node[align=center] at (9.5, -1.15) {\scriptsize Original image};
\node[align=center] at (9.5, -1.4) {\scriptsize prediction};

\node[align=left] at (10.75, -5.05) {$\rightarrow$};
\node[align=left] at (10.75, -2.45) {$=$};

\node[inner sep=0pt] (c) at (12, -5.05){\includegraphics[width=.1\textwidth]{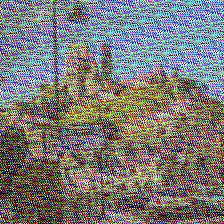}};
\node[align=left] at (12, -6.15) {\scriptsize Colliding};
\node[align=left] at (12, -6.4) {\scriptsize image};
\draw[->] (12, -4) -- ((12, -3.5);
\node[inner sep=0pt] (c) at (12, -2.45){\frame{\includegraphics[width=.1\textwidth]{IB_adv_ex/castle_logit.png}}};
\node[align=center] at (12, -1.15) {\scriptsize Colliding image};
\node[align=center] at (12, -1.4) {\scriptsize prediction};
\end{tikzpicture}

\caption{A colliding image created with the Null-space search. The perturbation is produced through the repetition of a randomly selected basis vector obtained from the kernel of the first trainable weight matrix of a ResNet-50. Both the original image and the colliding image have exactly the same output.}
\label{fig:ib-adv-gen}

\end{figure*}

\subsection{Finding data points with colliding features}
\label{finding_colliding_features}

To show the existence of colliding features, thus far, a variety of heuristic-based approaches were employed. Specifically, the method proposed in the work of \citet{sabour2015adversarial} allows for the creation of data points that come with similar predictions to other, dissimilar data points. This approach was used in the work of~\citet{jacobsen2018excessive} to create so-called invariance-based adversarial examples. These adversarial examples are constructed by minimizing an $\ell_p$ norm distance between two logit vectors, $g(\theta, \bm{x})$ and $g(\theta, \hat{\bm{x}})$, through the use of gradient descent: \mbox{$\hat{\bm{x}} = \min_{\hat{\bm{x}}} ||g(\theta, \bm{x}) - g(\theta, \hat{\bm{x}})||_{p}$}. The created invariance-based adversarial example $\hat{\bm{x}}$ now has similar logit predictions as $\bm{x}$. Independent from the aforementioned work, \citet{li2019approximate} also proposed another heuristic-based minimization method similar to the one given above that allows the creation of data points with colliding features.

Different from the approaches detailed above, we can take advantage of the observations made in Lemma~\ref{lem:linear_layer} and Remark~\ref{rem:conv_layer} and employ the basis vectors in the kernel of the trainable weights, which in turn would allow us to numerically create data points with colliding features. In order to demonstrate the feasibility of this approach, we define the Null-space search, a numerical method that allows for the creation of data points with colliding features.

\begin{definition}
\label{def:null-space-attack}
(Null-space search) Let $\bm{W}_1 \in \mathbb{R}^{q\times p}$ denote the weights of the first layer in a neural network for which $\nullity{(\bm{W}_1)}\geq1$. We define the Null-space search as $\psi_{\scriptscriptstyle{\bm{W}}}(\bm{x}, \bm{\varphi}) = \bm{x} + \beta \bm{\varphi}$, where $\beta$ is a scaling factor and where the perturbation vector $\bm{\varphi} \in \ker(\bm{W}_1)$ is one of the basis vectors selected from the kernel of $\bm{W}_1$, thus satisfying $\bm{\varphi} \in \mathbb{R}^{p} \,\,, \bm{W}_1 \bm{\varphi} = \bm{0}$, $\forall \bm{\varphi} \in \ker(\bm{W}_1)$.
\end{definition}

Data points created with the Null-space search not only have colliding features for the first weight but for all layers, including the final one (i.e., the prediction). Although this attack uses the basis vectors of the first weight for the sake of straightforward calculation, it is also possible to use the subsequent weights to create colliding data points (provided that they satisfy the conditions discussed in Lemma~\ref{lem:linear_layer} and Remark~\ref{rem:conv_layer}). However, in that scenario, one has to solve a system of non-linear equations to find the necessary modification.

In the rightmost column of the Table~\ref{tbl:class_seg_loc_nullities}, we provide the number of basis vectors in the kernel of the first trainable weight matrix of the evaluated models. As can be seen, most of the trained architectures indeed have $\nullity(\bm{W}_1) \geq 1$, thus allowing the Null-space search to create colliding data points. Note that having a single basis vector in the kernel of the first weight is sufficient to generate countless colliding data points with different linear combinations of $\bm{x}$ and $\bm{\varphi}$.

In Fig.~\ref{fig:ib-adv-gen}, we provide a visual illustration of the Null-space search for a classification scenario, through the usage of a trained ResNet-50 on ImageNet. As shown in Fig.~\ref{fig:ib-adv-gen}, the original image and its colliding counterpart have the same prediction, even though the adversarial example created with the proposed attack is visibly distorted by the perturbation obtained from the kernel of the first weight matrix. Let us now outline a number of properties of the Null-space search.

\textbf{Perturbation}\,\textendash\,Using the Null-space search when the first layer is a fully connected one is straightforward: we simply use a basis vector $\bm{\varphi}$ obtained from the kernel of the weight matrix. In the case when the first layer is a convolutional layer, we create the perturbation with $\bm{P} = [\bm{\varphi} \ldots \bm{\varphi}]^{\top}$. In both cases, the degree of perturbation can be controlled with $\beta$.

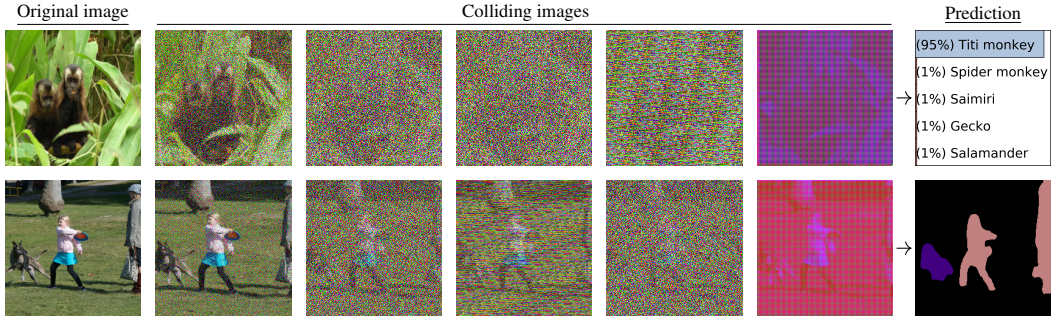
\begin{figure*}[t]
\centering
\begin{tikzpicture}
\centering

\node[align=center] at (0.2, 1.1) {\scriptsize \underline{Original image}};
\node[align=center] at (6.2, 1.1) {\scriptsize \underline{\phantom{--------------------------------------------------}Colliding images\phantom{-------------------------------------------------}}};
\node[align=center] at (12.3, 1.1) {\scriptsize \underline{Prediction}};

\def\sety1{0}
\node[inner sep=0pt] (a) at (0.2, \sety1)
    {\includegraphics[width=1.8cm]{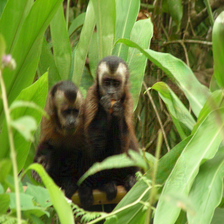}};
\node[inner sep=0pt] (a) at (2.2, \sety1)
    {\includegraphics[width=1.8cm]{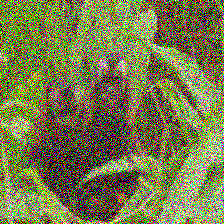}};
\node[inner sep=0pt] (c) at (4.2, \sety1)
    {\includegraphics[width=1.8cm]{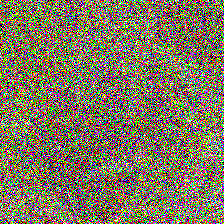}};
\node[inner sep=0pt] (c) at (6.2, \sety1)
    {\includegraphics[width=1.8cm]{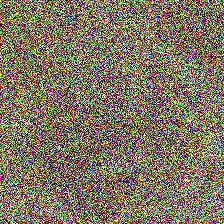}};
\node[inner sep=0pt] (c) at (8.2, \sety1)
    {\includegraphics[width=1.8cm]{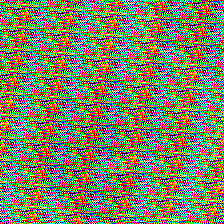}};
\node[inner sep=0pt] (c) at (10.2, \sety1)
    {\includegraphics[width=1.8cm]{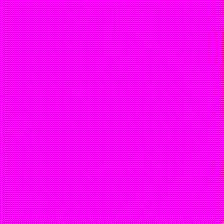}};
\draw[->] (11.15, \sety1) -- (11.35, \sety1);
\node[inner sep=0pt] (c) at (12.3, \sety1)    {\includegraphics[width=1.8cm]{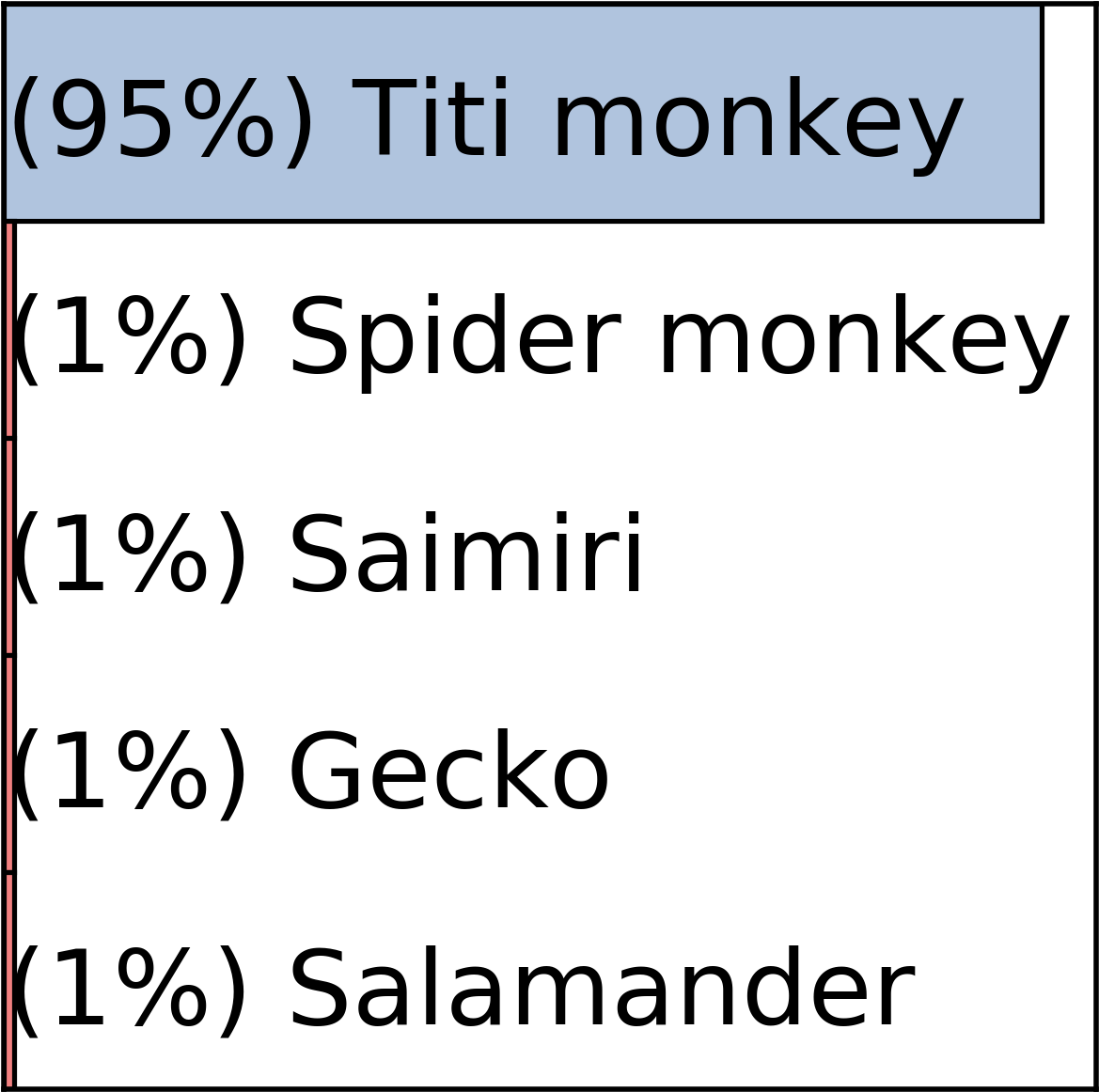}};
    
\def\sety1{-2}
\node[inner sep=0pt] (a) at (0.2, \sety1)
    {\includegraphics[width=1.8cm]{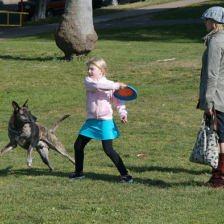}};
\node[inner sep=0pt] (a) at (2.2, \sety1)
    {\includegraphics[width=1.8cm]{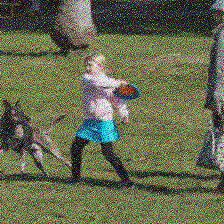}};
\node[inner sep=0pt] (c) at (4.2, \sety1)
    {\includegraphics[width=1.8cm]{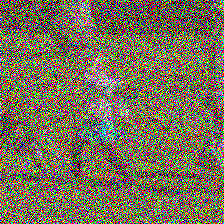}};
\node[inner sep=0pt] (c) at (6.2, \sety1)
    {\includegraphics[width=1.8cm]{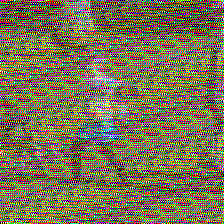}};
\node[inner sep=0pt] (c) at (8.2, \sety1)
    {\includegraphics[width=1.8cm]{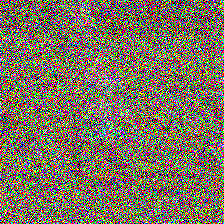}};
\node[inner sep=0pt] (c) at (10.2, \sety1)
    {\includegraphics[width=1.8cm]{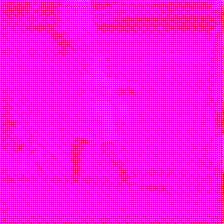}};
\draw[->] (11.15, \sety1) -- (11.35, \sety1);
\node[inner sep=0pt] (c) at (12.3, \sety1)
    {\includegraphics[width=1.8cm]{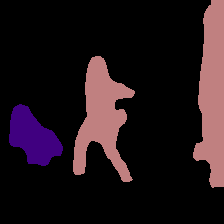}};
\end{tikzpicture}
\caption{Applications of the Null-space search for (top) classification and (bottom) segmentation. For classification and segmentation, we use Vit-B and MobileNet-V3, respectively. In the case of classification we provide the top-5 predictions and for segmentation, the predicted segmentation masks. Best viewed digitally.}
\label{fig:classification-segmentation-example}
\end{figure*}

In order to create the perturbation used in Fig.~\ref{fig:ib-adv-gen}, we relied on a single basis vector scaled with the same perturbation multiplier $\bm{P} = \beta \, [\bm{\varphi} \ldots \bm{\varphi}]^{\top}$, $\beta \in \mathbb{R}$, where this basis vector originates from the kernel of $\bm{W}_1$ (see the red rectangle). Although we used the same basis vector for producing the perturbation in Fig.~\ref{fig:ib-adv-gen}, we are not obliged to do so. Since any basis vector produces the same outcome (i.e., $(\bm{x} + \bm{\varphi}_i) \bm{W}_1 = \bm{x} \bm{W}_1$, $\forall \bm{\varphi}_i \in \ker(\bm{W}_1)$), it is possible to take different basis vectors (i.e., $\bm{P} = \beta \, [\bm{\varphi}_1 \, \bm{\varphi}_2\,   \ldots \, \bm{\varphi}_t]^{\top}$), thus producing different perturbations. Moreover, we are also not limited to the usage of a single perturbation multiplier. In this case, using individual perturbation multipliers (i.e., $\bm{P} = [\beta_1 \bm{\varphi}_1 \, \beta_2 \bm{\varphi}_2 \, \ldots \, \beta_t \bm{\varphi}_t]^{\top}$) makes it is possible to perturb a selected portion of the image while leaving other parts intact. In Fig.~\ref{fig:classification-segmentation-example}, we provide adversarial examples created with such approaches for a ViT-Base~\citep{dosovitskiy2021an} and MobileNet-V3~\citep{mobilenet_v3} trained on ImageNet.

\textbf{Task-agnostic approach}\,\textendash\,Note that the Null-space search is task-agnostic, meaning that, unlike other adversarial attacks that are proposed for a particular task (e.g., classification or segmentation), it is applicable for all types of neural networks built to solve any kind of problem. Note that the approach used for generating adversarial examples for segmentation, or any other task, is exactly the same as the above-described approach for classification.


\textbf{Creating adversarial examples}\,\textendash\,Because the Null-space search is a numerical method, it can also be used in conjunction with other adversarial attacks in order to create colliding adversarial examples. For this type of approach, instead of a genuine data point, an adversarial example generated by any other attack is used as an input for the Null-space search in the following way:
\begin{eqnarray}
\hat{\bm{x}} = \bm{x}' + \psi_{\scriptscriptstyle{\bm{W}}}(\bm{x}', \bm{\varphi})  \,,\,\,\,\,\, \bm{x}' = \Omega_{\epsilon}(\bm{x}),
\end{eqnarray}
where $\Omega_{\epsilon}$ represents any adversarial attack such as PGD~\citep{PGD_attack}. Even though the resulting adversarial example created with the Null-space search will have a different perturbation than the adversarial example created with the adversarial attack, they will have the same prediction $g(\theta, \hat{\bm{x}}) = g(\theta, \bm{x}')$. This outcome shows that it is indeed possible to have numerous adversarial examples that have the same, colliding features and prediction but different perturbations, thus indicating that the security flaws of DNNs with regards to various types of adversarial examples are far worse than what our intuition tells us.

\begin{figure*}[t]
\centering
\begin{tikzpicture}
\centering

\def\sety1{-5}
\node[align=center] at (-1, \sety1+0.9) {\scriptsize \underline{Original image}};
\node[align=center] at (0.8, \sety1+0.9) {\scriptsize \underline{Prediction}};
\node[align=center] at (3, \sety1+0.9) {\scriptsize \underline{$\epsilon$-adv. ex.}};
\node[align=center] at (4.8, \sety1+0.9) {\scriptsize \underline{Prediction}};
\node[align=center] at (9.2, \sety1+0.9) {\scriptsize \underline{\phantom{-----------}Colliding $\epsilon$-adversarial examples\phantom{-----------}}};

\node[inner sep=0pt] (a) at (-1, \sety1)
    {\includegraphics[width=1.5cm]{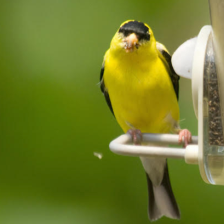}};
\draw[->] (-0.2, \sety1) -- (0, \sety1);
\node[inner sep=0pt] (a) at (0.8, \sety1)
    {\includegraphics[width=1.5cm]{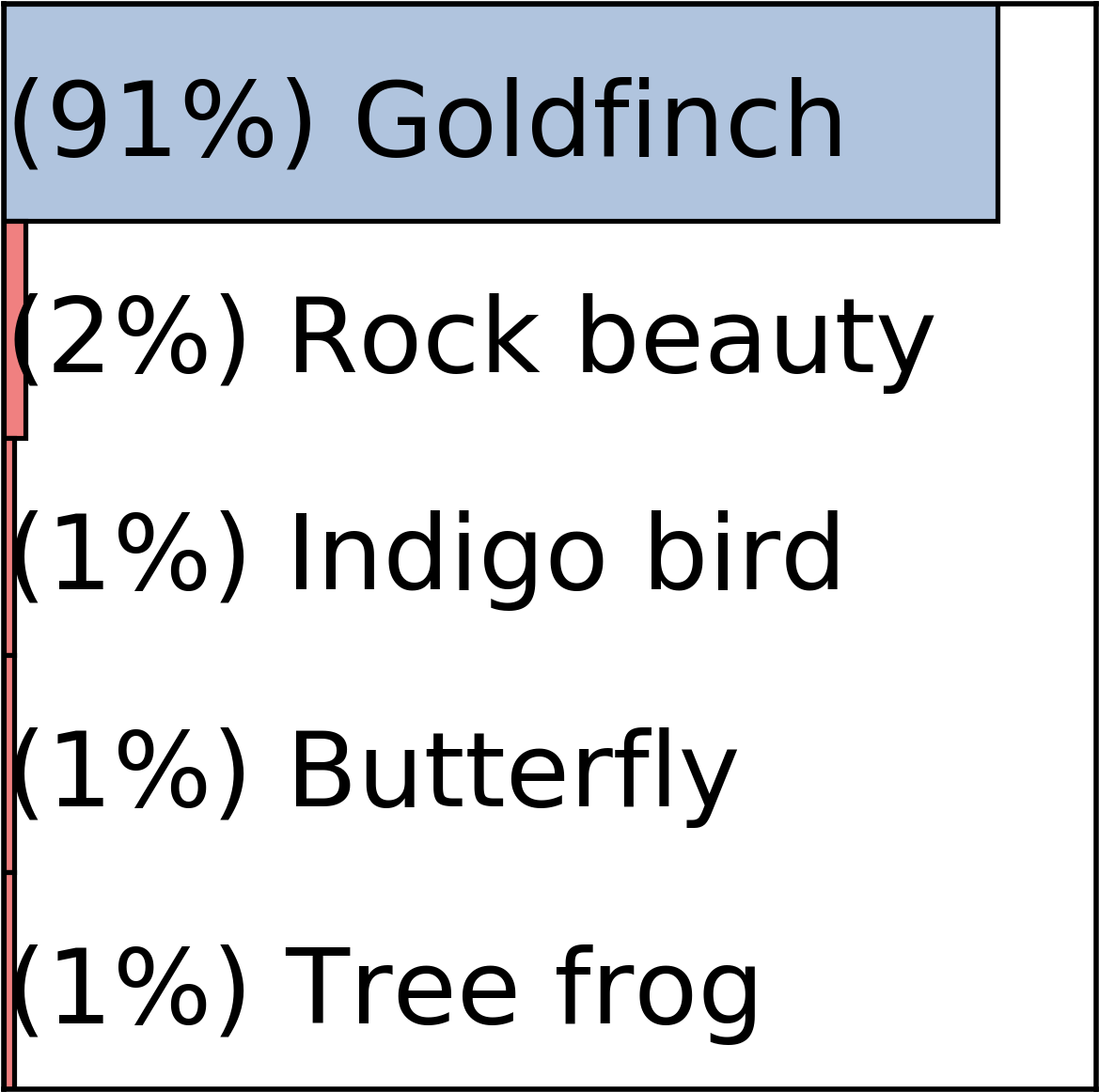}};

\node[inner sep=0pt] (b) at (3, \sety1)
    {\includegraphics[width=1.5cm]{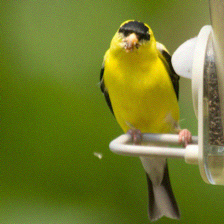}};
\draw[->] (3.8, \sety1) -- (4, \sety1);
\node[inner sep=0pt] (c) at (4.8, \sety1)
    {\includegraphics[width=1.5cm]{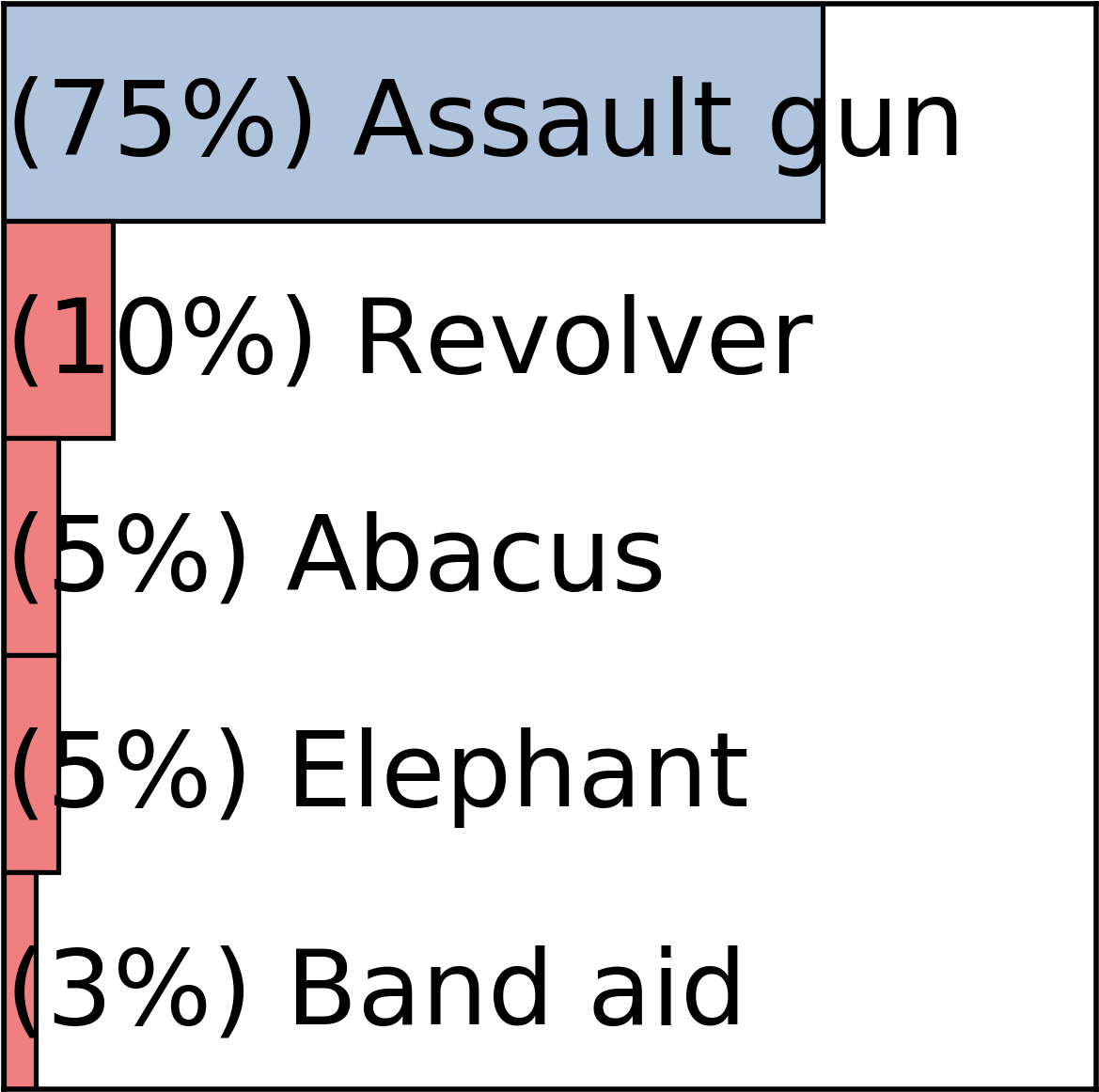}};

\draw[<-] (5.6, \sety1) -- (6, \sety1);
\node[inner sep=0pt] (c) at (6.8, \sety1)
    {\includegraphics[width=1.5cm]{cls_examples/bird_epsilon.png}};
\node[inner sep=0pt] (c) at (8.4, \sety1)
    {\includegraphics[width=1.5cm]{cls_examples/bird_epsilon.png}};
\node[inner sep=0pt] (c) at (10, \sety1)
    {\includegraphics[width=1.5cm]{cls_examples/bird_epsilon.png}};
\node[inner sep=0pt] (c) at (11.6, \sety1)
    {\includegraphics[width=1.5cm]{cls_examples/bird_epsilon.png}};
\def\sety1{-6.6}
\node[inner sep=0pt] (a) at (-1, \sety1)
    {\includegraphics[width=1.5cm]{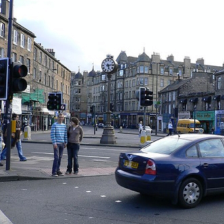}};
\draw[->] (-0.2, \sety1) -- (0, \sety1);
\node[inner sep=0pt] (a) at (0.8, \sety1)
    {\includegraphics[width=1.5cm]{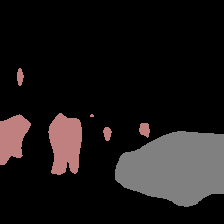}};

\node[inner sep=0pt] (b) at (3, \sety1)
    {\includegraphics[width=1.5cm]{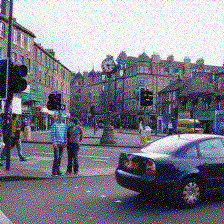}};
\draw[->] (3.8, \sety1) -- (4, \sety1);
\node[inner sep=0pt] (c) at (4.8, \sety1)
    {\includegraphics[width=1.5cm]{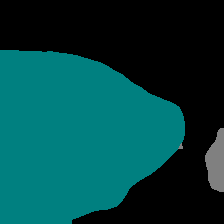}};

\draw[<-] (5.6, \sety1) -- (6, \sety1);
\node[inner sep=0pt] (c) at (6.8, \sety1)
    {\includegraphics[width=1.5cm]{Seg_examples/trafficreal_im.png}};
\node[inner sep=0pt] (c) at (8.4, \sety1)
    {\includegraphics[width=1.5cm]{Seg_examples/trafficreal_im.png}};
\node[inner sep=0pt] (c) at (10, \sety1)
    {\includegraphics[width=1.5cm]{Seg_examples/trafficreal_im.png}};
\node[inner sep=0pt] (c) at (11.6, \sety1)
    {\includegraphics[width=1.5cm]{Seg_examples/trafficreal_im.png}};

\end{tikzpicture}
\caption{Adversarial examples created with the Null-space search in conjunction with other adversarial attacks (PGD for classification and ASMA for segmentation). For classification and segmentation, we use Vit-B and MobileNet-V3, respectively. In the case of classification and segmentation, we provide the top-6 predictions and the predicted segmentation masks, respectively. Best viewed digitally with different levels of zoom.}
\label{fig:classification-segmentation-example_adv}
\end{figure*}

\section{Implications}\label{sec:implications}

We now discuss the implications of having data points with colliding features.

\textbf{Colliding predictions}\,\textendash\,The most straightforward consequence of data points with colliding features is that those data points will have the same prediction. This consequence, depending on the data points and the dataset at hand, can be seen as both a positive and a negative behavior of DNNs. In Figure~\ref{fig:classification-segmentation-example} we provide images with colliding features where the predictions of images on the same row are exactly the same. When the perturbation is mild and the object(s) of interest in those images are identifiable as their respective categories, colliding predictions may seem as behavior that is beneficial to DNNs. However, it is also possible to create data points with extremely large perturbation budgets where the objects of interest are not identifiable. In those cases, having a prediction that is the same as the original one becomes a detrimental behavior, since an extremely confident prediction for a particular category is made for an essentially unrecognizable image.

\textbf{Internal representations}\,\textendash\,Note that the internal representations (i.e., feature maps) of colliding images (for example, the ones created with the Null-space search), as well as the inputs used to create them, are the same, meaning that non-perturbation-based visualization approaches such as Grad-Cam~\citep{grad_cam} or Integrated Gradients~\citep{sundararajan2017axiomatic} will lead to the same visualization for differing images, further increasing the confusion with regards to the explainability of DNNs.

\textbf{Adversarial examples}\,\textendash\,Unlike the research on colliding images where two largely dissimilar images were shown to lead to the same features or prediction, the research on adversarial examples has shown that two images that seemingly look the same may have vastly different predictions. Our observations with regards to colliding images have an important relevance for adversarial examples, namely, that it is also possible to create colliding adversarial examples with the methods described in Section~\ref{Feature Collision}.

Although there is no unified formal definition of adversarial examples, the action space for adversaries is outlined in a number of previous research efforts~\citep{LBFGS,PGD_attack,motivating_rules_for_adv_research}. Based on the previous research work, for the classification system described in Section~\ref{Preliminaries}, an $\epsilon$-adversarial example, which is the most commonly used adversarial example, is defined as follows.

\begin{definition}
\label{def:adv_example}
($\epsilon$-adversarial examples) For a small perturbation $\Delta$ measured by an $\ell_p$-norm, $||\Delta||_p < \epsilon$, \mbox{$\hat{\bm{x}}_n = \bm{x}_n + \Delta \in \mathbb{R}^q$} is an $\epsilon$-adversarial example for a neural network with parameters $\theta$, created with the usage of an unperturbed image $\bm{x}_n$, if $G(\theta, \bm{x}_n) = y_n$ and $G(\theta, \hat{\bm{x}}_n) \neq y_n$.
\end{definition}

As described in Section~\ref{finding_colliding_features}, it is possible to employ the Null-space search with $\epsilon$-adversarial examples in order to generate even more adversarial examples with the same prediction. In order to demonstrate this, In Fig.~\ref{fig:classification-segmentation-example_adv}, we provide approaches where adversarial examples are created by combining the Null-space search with PGD~\citep{PGD_attack} or the Adaptive Segmentation Mask Attack (ASMA)~\citep{ozbulak2019impact} for classification and segmentation, respectively. In this scenario, colliding adversarial examples have the same prediction as the initial adversarial examples they are created from. Note that the degree of additive perturbation exercised by the Null-space search can be controlled with the selection of $\beta$ and differing basis vectors. In this case, added perturbation respects the initial $\ell_{\infty}$ limit set by the attacks, thus satisfying the property of invisible perturbation. This unsettling result indicates that the adversarial risk associated with DNNs is worse than previously thought, since every adversarial examples created with an attack may have countless colliding data points that are also adversarial examples.

\vspace{-1em}
\section{Where Theory Differs from Practice}\label{theory_practice}

In Section~\ref{Feature Collision} we laid out theoretical evidence for the existence of colliding features in DNNs from the perspective of weights. According to Lemma~\ref{lem:linear_layer}, by definition of basis vectors ($\bm{\varphi}$) in the nullity of matrices we have $\bm{\varphi}^{\top} \bm{W}_1 = \bm{0}$. As a result, any data point $\hat{\bm{x}} = \bm{x}+\bm{\varphi}$ obtained with $\bm{\varphi} \in \ker(\bm{W}_1)$ theoretically satisfies $\hat{\bm{x}}^{\top} \bm{W}_1 = \bm{x}^{\top} \bm{W}_1$, thus $\max (\hat{\bm{x}}^{\top} \bm{W}_1 - \bm{x}^{\top} \bm{W}_1) = 0$. However, in practice, for large matrices such as the ones in DNNs, $\bm{\varphi}^{\top} \bm{W}_1 = \bm{0}$ does not hold. Instead, $\bm{\varphi}^{\top} \bm{W}_1$ becomes approximately zero with $\bm{\varphi}^{\top} \bm{W}_1 \in [0, 10^{-7}]^{p}$ or $\bm{\varphi}^{\top} \bm{W}_1 \in [0, 10^{-15}]^{p}$, depending on the numerical precision of the operation (i.e., single or double).

As a result of the above observation, $\max (\hat{\bm{x}}^{\top} \bm{W}_1 - \bm{x}^{\top} \bm{W}_1) = 0$, more often than not, also does not hold. Although this difference is negligible for the output of the first layer, proceeding operations in each consecutive layer may exacerbate this difference. As a consequence of this small difference in the output of the first layer, two data points that are meant to be theoretically identical in terms of output (i.e., final layer) may be similar but not (exactly) the same. Consequently, targeting later layers of DNNs would naturally yield predictions that are more stable in terms of similarity of the output but in such cases, a system of non-linear equations would have to be solved to find an exact perturbation. We leave this analysis on the stability of perturbations to future studies.

To sum up, although theoretical evidence may hint at certain types of results, shortcomings associated with the numerical precision of computation may result in different experimental observations. We hope that this discussion regarding the probable difference between theoretical and empirical results can alleviate any confusion that might arise in future efforts in this line of research.

\section{Conclusions and Outlook}
\label{Conclusions and Outlook}

In this paper, we characterized colliding features using a formal approach, investigating the characteristics of DNNs to such data points that may come with similar, colliding features. Through extensive formal analysis of the trainable weights, which are arguably the most important building blocks of DNNs, we identified the conditions for which colliding data points are inevitable. Based on the knowledge obtained from our formal analysis, we proposed the Null-space search, a numerical method that can create colliding data points without relying on heuristics. Linking our research to the recently discovered vulnerability of adversarial examples, we also showed the potential threat of having numerous adversarial examples due to this shortcoming of colliding features.

Although research on knowledge distillation is receiving an increasing amount of attention nowadays, the pursuit of state-of-the-art results in recent years led to novel architectures that are not only deeper but also wider. As we have discussed in Section~\ref{Feature Collision}, having a larger neural network also increases the chance that the network is vulnerable to adversarial examples. Supporting the ideas proposed by~\citet{heo2019knowledge}, and \citet{goldblum2020adversarially}, we acknowledge that knowledge distillation and having compact networks are a necessity in order to minimize the adversarial risk.

In future work, we believe the Null-space search, as well as the metrics evaluated in Table~\ref{tbl:class_seg_loc_nullities}, could be used as benchmarking methods to quantify and compare the adversarial risk associated with models. Specifically, we would like to analyze whether or not, for example, $\frac{1}{n_{\theta}} \nu(\theta)$, $\frac{1}{n_{\theta}} \mu (\theta)$, and $\sum_{\bm{W}_i \in \theta}  \nullity(\bm{W}_i)$ could be used as metrics to investigate the vulnerability of models to adversarial examples. Given the lack of easily available quantitative evaluation metrics in the field of adversarial attacks, if such methods show correlation with the easiness of creation of adversarial examples, they can be expected to be helpful for fairly evaluating the adversarial risk associated with models.

\bibliographystyle{abbrvnat}
\bibliography{20_21}

\begin{thebibliography}{49}
\providecommand{\natexlab}[1]{#1}
\providecommand{\url}[1]{\texttt{#1}}
\expandafter\ifx\csname urlstyle\endcsname\relax
  \providecommand{\doi}[1]{doi: #1}\else
  \providecommand{\doi}{doi: \begingroup \urlstyle{rm}\Url}\fi

\bibitem[Biggio et~al.(2013)Biggio, Corona, Maiorca, Nelson, {\v{S}}rndi{\'c},
  Laskov, Giacinto, and Roli]{biggio2013evasion}
B.~Biggio, I.~Corona, D.~Maiorca, B.~Nelson, N.~{\v{S}}rndi{\'c}, P.~Laskov,
  G.~Giacinto, and F.~Roli.
\newblock {Evasion Attacks Against Machine Learning At Test Time}.
\newblock In \emph{Joint European Conference on Machine Learning and Knowledge
  Discovery in Databases}, 2013.

\bibitem[Carlini and Wagner(2017)]{CW_Attack}
N.~Carlini and D.~A. Wagner.
\newblock {Towards Evaluating The Robustness of Neural Networks}.
\newblock \emph{2017 IEEE Symposium on Security and Privacy}, 2017.

\bibitem[Chaturvedi and Garain(2020)]{mimic_fool}
A.~Chaturvedi and U.~Garain.
\newblock {Mimic and Fool: A Task-Agnostic Adversarial Attack}.
\newblock \emph{IEEE Transactions on Neural Networks and Learning Systems},
  2020.

\bibitem[Chernikova et~al.(2019)Chernikova, Oprea, Nita-Rotaru, and
  Kim]{self_driving_adversarial}
A.~Chernikova, A.~Oprea, C.~Nita-Rotaru, and B.~Kim.
\newblock {Are Self-Driving Cars Secure? Evasion Attacks Against Deep Neural
  Networks For Steering Angle Prediction}.
\newblock \emph{IEEE Security and Privacy Workshops}, 2019.

\bibitem[Devlin et~al.(2018)Devlin, Chang, Lee, and Toutanova]{devlin2018bert}
J.~Devlin, M.-W. Chang, K.~Lee, and K.~Toutanova.
\newblock Bert: Pre-training of deep bidirectional transformers for language
  understanding.
\newblock \emph{arXiv preprint arXiv:1810.04805}, 2018.

\bibitem[Dong et~al.(2018)Dong, Liao, Pang, Su, Zhu, Hu, and
  Li]{dong2018boosting}
Y.~Dong, F.~Liao, T.~Pang, H.~Su, J.~Zhu, X.~Hu, and J.~Li.
\newblock {Boosting Adversarial Attacks with Momentum}.
\newblock In \emph{Proceedings of the IEEE conference on computer vision and
  pattern recognition}, 2018.

\bibitem[Dosovitskiy et~al.(2021)Dosovitskiy, Beyer, Kolesnikov, Weissenborn,
  Zhai, Unterthiner, Dehghani, Minderer, Heigold, Gelly, Uszkoreit, and
  Houlsby]{dosovitskiy2021an}
A.~Dosovitskiy, L.~Beyer, A.~Kolesnikov, D.~Weissenborn, X.~Zhai,
  T.~Unterthiner, M.~Dehghani, M.~Minderer, G.~Heigold, S.~Gelly, J.~Uszkoreit,
  and N.~Houlsby.
\newblock {An Image is Worth 16x16 Words: Transformers for Image Recognition at
  Scale}.
\newblock In \emph{International Conference on Learning Representations}, 2021.

\bibitem[Gilmer et~al.(2018)Gilmer, Adams, Goodfellow, Andersen, and
  Dahl]{motivating_rules_for_adv_research}
J.~Gilmer, R.~P. Adams, I.~Goodfellow, D.~Andersen, and G.~E. Dahl.
\newblock {Motivating The Rules Of The Game For Adversarial Example Research}.
\newblock \emph{CoRR}, abs/1807.06732, 2018.

\bibitem[Glorot et~al.(2011)Glorot, Bordes, and Bengio]{ReLU}
X.~Glorot, A.~Bordes, and Y.~Bengio.
\newblock {Deep Sparse Rectifier Neural Networks}.
\newblock In \emph{International Conference on Artificial Intelligence and
  Statistics}, 2011.

\bibitem[Goldblum et~al.(2020)Goldblum, Fowl, Feizi, and
  Goldstein]{goldblum2020adversarially}
M.~Goldblum, L.~Fowl, S.~Feizi, and T.~Goldstein.
\newblock {Adversarially Robust Distillation}.
\newblock In \emph{Proceedings of the AAAI Conference on Artificial
  Intelligence}, 2020.

\bibitem[Goodfellow et~al.(2015)Goodfellow, Shlens, and
  Szegedy]{Goodfellow-expharnessing}
I.~Goodfellow, J.~Shlens, and C.~Szegedy.
\newblock {Explaining and Harnessing Adversarial Examples}.
\newblock \emph{International Conference on Learning Representations}, 2015.

\bibitem[He et~al.(2016)He, Zhang, Ren, and Sun]{resnet}
K.~He, X.~Zhang, S.~Ren, and J.~Sun.
\newblock {Deep Residual Learning For Image Recognition}.
\newblock In \emph{Proceedings of the IEEE Conference on Computer Vision and
  Pattern Recognition}, 2016.

\bibitem[He et~al.(2017)He, Gkioxari, Doll{\'a}r, and Girshick]{mask_rcnn}
K.~He, G.~Gkioxari, P.~Doll{\'a}r, and R.~Girshick.
\newblock {Mask R-CNN}.
\newblock In \emph{Proceedings of the IEEE International Conference on Computer
  Vision}, 2017.

\bibitem[Hein et~al.(2019)Hein, Andriushchenko, and Bitterwolf]{hein2019relu}
M.~Hein, M.~Andriushchenko, and J.~Bitterwolf.
\newblock {Why ReLU Networks Yield High-confidence Predictions Far Away From
  the Training Data and How to Mitigate the Problem}.
\newblock In \emph{Proceedings of the IEEE Conference on Computer Vision and
  Pattern Recognition}, 2019.

\bibitem[Hendrycks and Gimpel(2016)]{hendrycks2016gelu}
D.~Hendrycks and K.~Gimpel.
\newblock {Gaussian Error Linear Units (GELUs)}.
\newblock \emph{CoRR}, abs/1606.08415, 2016.

\bibitem[Heo et~al.(2019)Heo, Lee, Yun, and Choi]{heo2019knowledge}
B.~Heo, M.~Lee, S.~Yun, and J.~Y. Choi.
\newblock {Knowledge Distillation with Adversarial Samples Supporting Decision
  Boundary}.
\newblock In \emph{Proceedings of the AAAI Conference on Artificial
  Intelligence}, 2019.

\bibitem[Horn and Johnson(2013)]{horn2012matrix}
R.~A. Horn and C.~R. Johnson.
\newblock \emph{{Matrix Analysis}}.
\newblock Cambridge university press, 2013.

\bibitem[Hosseini et~al.(2019)Hosseini, Kannan, and
  Poovendran]{hosseini2019are_odds_odd}
H.~Hosseini, S.~Kannan, and R.~Poovendran.
\newblock {Are Odds Really Odd? Bypassing Statistical Detection Of Adversarial
  Examples}.
\newblock \emph{CoRR}, abs/1907.12138, 2019.

\bibitem[Howard et~al.(2019)Howard, Sandler, Chu, Chen, Chen, Tan, Wang, Zhu,
  Pang, Vasudevan, et~al.]{mobilenet_v3}
A.~Howard, M.~Sandler, G.~Chu, L.-C. Chen, B.~Chen, M.~Tan, W.~Wang, Y.~Zhu,
  R.~Pang, V.~Vasudevan, et~al.
\newblock {Searching for Mobilenetv3}.
\newblock In \emph{Proceedings of the IEEE International Conference on Computer
  Vision}, 2019.

\bibitem[Huang et~al.(2017)Huang, Liu, Van Der~Maaten, and
  Weinberger]{densenet}
G.~Huang, Z.~Liu, L.~Van Der~Maaten, and K.~Q. Weinberger.
\newblock {Densely Connected Convolutional Networks}.
\newblock In \emph{Proceedings of the IEEE Conference on Computer Vision and
  Pattern Recognition}, 2017.

\bibitem[Iandola et~al.(2016)Iandola, Han, Moskewicz, Ashraf, Dally, and
  Keutzer]{squeezenet}
F.~N. Iandola, S.~Han, M.~W. Moskewicz, K.~Ashraf, W.~J. Dally, and K.~Keutzer.
\newblock Squeezenet: Alexnet-level accuracy with 50x fewer parameters and< 0.5
  mb model size.
\newblock \emph{CoRR}, abs/1602.07360, 2016.

\bibitem[Ilyas et~al.(2019)Ilyas, Santurkar, Tsipras, Engstrom, Tran, and
  Madry]{ilyas2019adversarial}
A.~Ilyas, S.~Santurkar, D.~Tsipras, L.~Engstrom, B.~Tran, and A.~Madry.
\newblock {Adversarial Examples Are Not Bugs, They Are Features}.
\newblock In \emph{Advances in Neural Information Processing Systems}, pages
  125--136, 2019.

\bibitem[Jacobsen et~al.(2019)Jacobsen, Behrmann, Zemel, and
  Bethge]{jacobsen2018excessive}
J.-H. Jacobsen, J.~Behrmann, R.~Zemel, and M.~Bethge.
\newblock {Excessive Invariance Causes Adversarial Vulnerability}.
\newblock In \emph{International Conference on Learning Representations}, 2019.

\bibitem[Krizhevsky et~al.(2012)Krizhevsky, Sutskever, and Hinton]{Alexnet}
A.~Krizhevsky, I.~Sutskever, and G.~E. Hinton.
\newblock {ImageNet Classification with Deep Convolutional Neural Networks}.
\newblock In \emph{Advances in Neural Information Processing Systems}, 2012.

\bibitem[Kurakin et~al.(2016)Kurakin, Goodfellow, and Bengio]{IFGS}
A.~Kurakin, I.~Goodfellow, and S.~Bengio.
\newblock {Adversarial Examples In The Physical World}.
\newblock \emph{Workshop Track, International Conference on Learning
  Representations}, 2016.

\bibitem[LeCun et~al.(1998)LeCun, Bottou, Bengio, and
  Haffner]{lecun1998gradient}
Y.~LeCun, L.~Bottou, Y.~Bengio, and P.~Haffner.
\newblock {Gradient-Based Learning Applied To Document Recognition}.
\newblock \emph{Proceedings of the IEEE}, 1998.

\bibitem[Li et~al.(2019)Li, Zhang, and Malik]{li2019approximate}
K.~Li, T.~Zhang, and J.~Malik.
\newblock {Approximate Feature Collisions in Neural Nets}.
\newblock \emph{Advances in Neural Information Processing Systems}, 32, 2019.

\bibitem[Lin et~al.(2017)Lin, Goyal, Girshick, He, and
  Doll{\'a}r]{focal_loss_retina_net}
T.-Y. Lin, P.~Goyal, R.~Girshick, K.~He, and P.~Doll{\'a}r.
\newblock {Focal Loss for Dense Object Detection}.
\newblock In \emph{Proceedings of the IEEE International Conference on Computer
  Vision}, 2017.

\bibitem[Long et~al.(2015)Long, Shelhamer, and Darrell]{Fcn8s}
J.~Long, E.~Shelhamer, and T.~Darrell.
\newblock {Fully Convolutional Networks For Semantic Segmentation}.
\newblock In \emph{Proceedings of the IEEE Conference on Computer Vision and
  Pattern Recognition}, pages 3431--3440, 2015.

\bibitem[Madry et~al.(2018)Madry, Makelov, Schmidt, Tsipras, and
  Vladu]{PGD_attack}
A.~Madry, A.~Makelov, L.~Schmidt, D.~Tsipras, and A.~Vladu.
\newblock {Towards Deep Learning Models Resistant To Adversarial Attacks}.
\newblock \emph{International Conference on Learning Representations}, 2018.

\bibitem[Ozbulak et~al.(2019)Ozbulak, Van~Messem, and
  De~Neve]{ozbulak2019impact}
U.~Ozbulak, A.~Van~Messem, and W.~De~Neve.
\newblock {Impact Of Adversarial Examples On Deep Learning Models For
  Biomedical Image Segmentation}.
\newblock In \emph{International Conference on Medical Image Computing and
  Computer-Assisted Intervention}, 2019.

\bibitem[Ren et~al.(2015)Ren, He, Girshick, and Sun]{faster_rcnn}
S.~Ren, K.~He, R.~Girshick, and J.~Sun.
\newblock {Faster R-CNN: Towards Real-Time Object Detection with Region
  Proposal Networks}.
\newblock In \emph{Advances in Neural Information Processing Systems}, 2015.

\bibitem[Ronneberger et~al.(2015)Ronneberger, Fischer, and Brox]{Unet}
O.~Ronneberger, P.~Fischer, and T.~Brox.
\newblock {U-Net: Convolutional Networks For Biomedical Image Segmentation}.
\newblock In N.~Navab, J.~Hornegger, W.~M. Wells, and A.~F. Frangi, editors,
  \emph{Medical Image Computing and Computer-Assisted Intervention}, 2015.

\bibitem[Sabour et~al.(2016)Sabour, Cao, Faghri, and
  Fleet]{sabour2015adversarial}
S.~Sabour, Y.~Cao, F.~Faghri, and D.~J. Fleet.
\newblock {Adversarial Manipulation Of Deep Representations}.
\newblock In \emph{International Conference on Learning Representations}, 2016.

\bibitem[Selvaraju et~al.(2016)Selvaraju, Das, Vedantam, Cogswell, Parikh, and
  Batra]{grad_cam}
R.~R. Selvaraju, A.~Das, R.~Vedantam, M.~Cogswell, D.~Parikh, and D.~Batra.
\newblock {Grad-Cam: Why Did You Say That? Visual Explanations From Deep
  Networks Via Gradient-Based Localization}.
\newblock \emph{Proceedings of the IEEE Conference on Computer Vision and
  Pattern Recognition}, 2016.

\bibitem[Simonyan and Zisserman(2015)]{VGG}
K.~Simonyan and A.~Zisserman.
\newblock {Very Deep Convolutional Networks For Large-Scale Image Recognition}.
\newblock \emph{International Conference on Learning Representations}, 2015.

\bibitem[Springenberg et~al.(2014)Springenberg, Dosovitskiy, Brox, and
  Riedmiller]{guided_backprop}
J.~T. Springenberg, A.~Dosovitskiy, T.~Brox, and M.~Riedmiller.
\newblock {Striving For Simplicity: The All Convolutional Net}.
\newblock \emph{CoRR}, abs/1412.6806, 2014.

\bibitem[Subramanya et~al.(2019)Subramanya, Pillai, and
  Pirsiavash]{subramanya2019fooling}
A.~Subramanya, V.~Pillai, and H.~Pirsiavash.
\newblock Fooling network interpretation in image classification.
\newblock In \emph{Proceedings of the IEEE/CVF International Conference on
  Computer Vision}, 2019.

\bibitem[Sundararajan et~al.(2017)Sundararajan, Taly, and
  Yan]{sundararajan2017axiomatic}
M.~Sundararajan, A.~Taly, and Q.~Yan.
\newblock {Axiomatic Attribution for Deep Networks}.
\newblock In \emph{International conference on machine learning}. PMLR, 2017.

\bibitem[Szegedy et~al.(2014)Szegedy, Zaremba, Sutskever, Bruna, Erhan,
  Goodfellow, and Fergus]{LBFGS}
C.~Szegedy, W.~Zaremba, I.~Sutskever, J.~Bruna, D.~Erhan, I.~Goodfellow, and
  R.~Fergus.
\newblock {Intriguing Properties Of Neural Networks}.
\newblock \emph{International Conference on Learning Representations}, 2014.

\bibitem[Szegedy et~al.(2016)Szegedy, Vanhoucke, Ioffe, Shlens, and
  Wojna]{inceptionv3}
C.~Szegedy, V.~Vanhoucke, S.~Ioffe, J.~Shlens, and Z.~Wojna.
\newblock {Rethinking The Inception Architecture For Computer Vision}.
\newblock In \emph{Proceedings of the IEEE Conference on Computer Vision and
  Pattern Recognition}, 2016.

\bibitem[Touvron et~al.(2020)Touvron, Cord, Douze, Massa, Sablayrolles, and
  J{\'e}gou]{touvron2020training}
H.~Touvron, M.~Cord, M.~Douze, F.~Massa, A.~Sablayrolles, and H.~J{\'e}gou.
\newblock {Training Data-efficient Image Transformers \& Distillation Through
  Attention}.
\newblock \emph{CoRR}, abs/2012.12877, 2020.

\bibitem[Tram{\`e}r et~al.(2020)Tram{\`e}r, Behrmann, Carlini, Papernot, and
  Jacobsen]{tramer2020fundamental}
F.~Tram{\`e}r, J.~Behrmann, N.~Carlini, N.~Papernot, and J.-H. Jacobsen.
\newblock Fundamental tradeoffs between invariance and sensitivity to
  adversarial perturbations.
\newblock In \emph{International Conference on Machine Learning}, pages
  9561--9571. PMLR, 2020.

\bibitem[Tramer et~al.(2020)Tramer, Carlini, Brendel, and
  Madry]{CarliniTeam_2020_defense_evaluation}
F.~Tramer, N.~Carlini, W.~Brendel, and A.~Madry.
\newblock {On Adaptive Attacks to Adversarial Example Defenses}.
\newblock \emph{Advances in Neural Information Processing Systems}, 2020.

\bibitem[Vandersmissen et~al.(2019)Vandersmissen, Knudde, Jalalvand, Couckuyt,
  Dhaene, and De~Neve]{vandersmissen2019indoor}
B.~Vandersmissen, N.~Knudde, A.~Jalalvand, I.~Couckuyt, T.~Dhaene, and
  W.~De~Neve.
\newblock {Indoor Human Activity Recognition Using High-Dimensional Sensors And
  Deep Neural Networks}.
\newblock \emph{Neural Computing and Applications}, pages 1--15, 2019.

\bibitem[Vaswani et~al.(2017)Vaswani, Shazeer, Parmar, Uszkoreit, Jones, Gomez,
  Kaiser, and Polosukhin]{attention}
A.~Vaswani, N.~Shazeer, N.~Parmar, J.~Uszkoreit, L.~Jones, A.~N. Gomez, L.~u.
  Kaiser, and I.~Polosukhin.
\newblock {Attention is All you Need}.
\newblock In \emph{{Advances in Neural Information Processing Systems}}, 2017.

\bibitem[Wang et~al.(2020)Wang, Wang, Li, Zhang, and
  Lin]{wang2020transferable_cross}
H.~Wang, G.~Wang, Y.~Li, D.~Zhang, and L.~Lin.
\newblock {Transferable, Controllable, and Inconspicuous Adversarial Attacks on
  Person Re-identification With Deep Mis-Ranking}.
\newblock In \emph{Proceedings of the IEEE Conference on Computer Vision and
  Pattern Recognition}, pages 342--351, 2020.

\bibitem[Zhao et~al.(2017)Zhao, Shi, Qi, Wang, and Jia]{pspnet}
H.~Zhao, J.~Shi, X.~Qi, X.~Wang, and J.~Jia.
\newblock {Pyramid Scene Parsing Network}.
\newblock In \emph{Proceedings of the IEEE Conference on Computer Vision and
  Pattern Recognition}, 2017.

\bibitem[Zuallaert et~al.(2018)Zuallaert, Godin, Kim, Soete, Saeys, and
  De~Neve]{zuallaert2018splicerover}
J.~Zuallaert, F.~Godin, M.~Kim, A.~Soete, Y.~Saeys, and W.~De~Neve.
\newblock {SpliceRover: Interpretable Convolutional Neural Networks for
  Improved Splice Site Prediction}.
\newblock \emph{Bioinformatics}, 2018.

\end{thebibliography}

\end{document}